\documentclass[letterpaper, 10 pt, conference]{ieeeconf}

\IEEEoverridecommandlockouts                              

\usepackage{graphics} 
\usepackage{epsfig} 
\usepackage{times} 
\usepackage{amsmath} 
\usepackage{amssymb}  
\usepackage{amsmath, bm}
\usepackage{float}
\usepackage{subfigure}
\usepackage{threeparttable}
\usepackage{tabularx}
\usepackage{multirow}
\usepackage{booktabs}
\usepackage[linesnumbered,ruled,vlined]{algorithm2e}
\usepackage{algorithmicx}
\usepackage{pifont}
\usepackage{enumerate}
\usepackage[breaklinks=true]{hyperref}
\usepackage{color}
\newcolumntype{Y}{>{\centering\arraybackslash}m{0.7cm}}

\newcommand{\magenta}[1]{\textcolor{magenta}{#1}}
\title{\LARGE \bf
A Closed-Loop Bin Picking System for Entangled Wire Harnesses\\using Bimanual and Dynamic Manipulation
}

\author{Xinyi Zhang$^{1*}$, 
Yukiyasu Domae$^{2}$, Weiwei Wan$^{1}$ and Kensuke Harada$^{1,2}$
\thanks{*Correspond to: {\tt\small xinyiz0931@gmail.com}}
\thanks{$^{1}$Graduate School of Engineering Science, Osaka University, Japan}%
\thanks{$^{2}$Industrial Cyber Physical Systems Research Center, National Institute of Advanced Industrial Science and Technology (AIST), Japan}
}

\begin{document}

\maketitle
\thispagestyle{empty}
\pagestyle{empty}

\begin{abstract}
    This paper addresses the challenge of industrial bin picking using entangled wire harnesses. Wire harnesses are essential in manufacturing but poses challenges in automation due to their complex geometries and propensity for entanglement. Our previous work tackled this issue by proposing a quasi-static pulling motion to separate the entangled wire harnesses. However, it still lacks sufficiency and generalization to various shapes and structures. In this paper, we deploy a dual-arm robot that can grasp, extract and disentangle wire harnesses from dense clutter using dynamic manipulation. The robot can swing to dynamically discard the entangled objects and regrasp to adjust the undesirable grasp pose. To improve the robustness and accuracy of the system, we leverage a closed-loop framework that uses haptic feedback to detect entanglement in real-time and flexibly adjust system parameters. Our bin picking system achieves an overall success rate of 91.2\% in the real-world experiments using two different types of long wire harnesses. It demonstrates the effectiveness of our system in handling various wire harnesses for industrial bin picking. The project’s website can be found at \magenta{\url{https://xinyiz0931.github.io/dynamic}}. 
\end{abstract}


\section{Introduction}
\label{sec:intro}

    Robotic bin picking has been developed for decades to automate the manufacturing process. It enables a robot to pick objects randomly placed in a bin and provide them to the assembly process. Utilizing bin picking in the assembly line can eliminate the need for preparing parts feeders and relieve human workers from tedious and repetitive workload. Existing bin picking approaches always focus on picking rigid objects \cite{kirkegaard2006bin,liu2012fast,buchholz2013efficient,domae2014fast,harada2016initial,matsumura2018learning,tachikake2020learning} while deformable objects pose new challenges. Wire harnesses are crucial components in the assembly of electric drive products. Due to their complex structures, which comprise both deformable cables and rigid components, wire harnesses in a cluttered environment are prone to entanglement. Developing bin picking system for entangled wire harnesses is extremely difficult. Object recognition and grasp detection becomes challenging in such complex scenarios involving with rich contact and environmental uncertainties. In the case of the robot grasping the end of the objects, executing disentangling motions becomes insufficient. Simulated training or obtaining models for wire harnesses still remains an open problem while training in the real world is time-consuming. Additionally, the manipulatable range of the robot in a standard bin picking cell has limited the maximum length of wire harnesses that can be handled. These difficulties led manufacturing industries to rely on human workers to manually separate entangled wire harnesses in the assembly processes. 
    
    In our previous work \cite{zhang2022learning}, we learned a sequential policy that performs a circle-drawing trajectory to disentangle the wire harnesses. However, as the number of objects in a bin increases or when adapting to unseen objects types, the patterns of entanglement become unpredictable, making visual recognition and the circling motion insufficient. Moreover, wire harnesses often exceed the robot's reachable range, further diminishing the performance of quasi-static motion primitives for disentangling them. Therefore, more effective motion primitives and multiple sensing capabilities are highly demanded to ensure a robust, accurate and versatile bin picking system for wire harnesses.


    
    To further address these challenges, we present a bin picking system for entangled wire harnesses with the following key components: 

\begin{itemize}
    \item We propose two motion primitives: swing and regrasping, specifically designed for disentangling of long wire harnesses. The swing motion with a high velocity can dynamically extract the target from the clutter. On the other hand, regrasping enables the robot to grasp the target at its middle section, creating sufficient space for disentangling process. 
    \item We present a closed-loop system that utilizes haptic feedback to detect entanglement in real-time and tunes the system parameters online. Unlike open-loop policies without error recovery, our system closes the loop by incorporating force feedback, enhancing the robustness and efficiency for picking entangled wire harnesses. 
\end{itemize}

\begin{figure*}[t] 
    \centering
    \includegraphics[width=\linewidth]{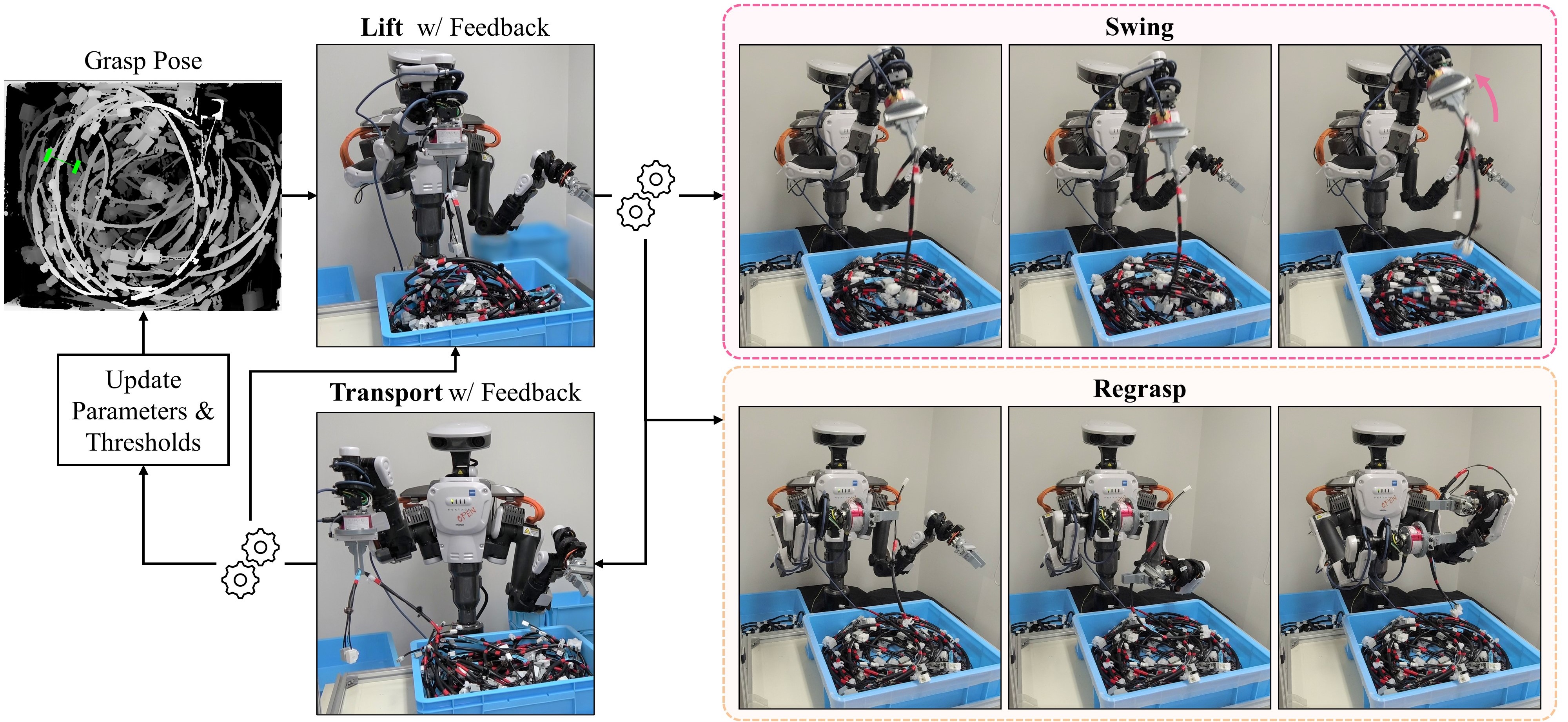}    
    \caption{Overall process of picking entangled wire harnesses. }
    \label{fig:overview} 
\end{figure*}
    
    Our primary contribution is a unique bin picking system for wire harnesses that leverages dynamic and bimanual manipulation as disentangling strategies. We propose a haptic-guided closed-loop algorithm with failure detection and recovery in real-time. Real-world experiments suggest our policy can significantly improve the average success rates compared with priors works.

\section{Related Work}

\subsection{Industrial Bin Picking}

    The classic approach to industrial bin picking matches known object models to the scene and locates the objects \cite{liu2012fast,choi2012voting,yang2021probabilistic,liu20216d} and plan force closure grasps for the target object \cite{dupuis2008two,buchholz2014combining,harada2013probabilistic}. These analytical approaches provide higher robustness and accuracy and have already been successfully implemented in manufacturing. Furthermore, alternative approaches that do not rely on object models can also be adapted for practical bin picking processes. Domae et al. \cite{domae2014fast} proposed a method to plan grasps by considering collisions between the gripper and objects using a single depth image. Recently, deep learning exhibites great potential in addressing the remaining challenges in bin picking, including enhancing the grasp quality \cite{mahler2017dex,matsumura2018learning} and picking difficult objects \cite{tachikake2020learning,tong2021dig,morino2020sheet,ishige2020blind}. While considerable progress has been made with rigid tangled-prone objects \cite{matsumura2019learning,zhang2021topological,zhang2023learning}, deformable objects with complex structures, such as wire harnesses, remain relatively unexplored. Several works have addressed factory automation problems related to wire harnesses \cite{guo2022visual,jiang2011robotized,zhou2020practical}. Zhang et al. \cite{zhang2022learning} first tackled the problem of bin picking for wire harnesses by proposing a pulling motion and learning an sequential policy. However, the performance decreases when the bin contains a large number of objects or when adapting to longer wire harnesses. In this paper, we propose a hybrid system that combines an analytical algorithm with a learned vision module. This system can achieve higher success rate and robustness in picking various wire harnesses. 
    
\begin{figure*}[t] 
    \centering
    \includegraphics[width=\linewidth]{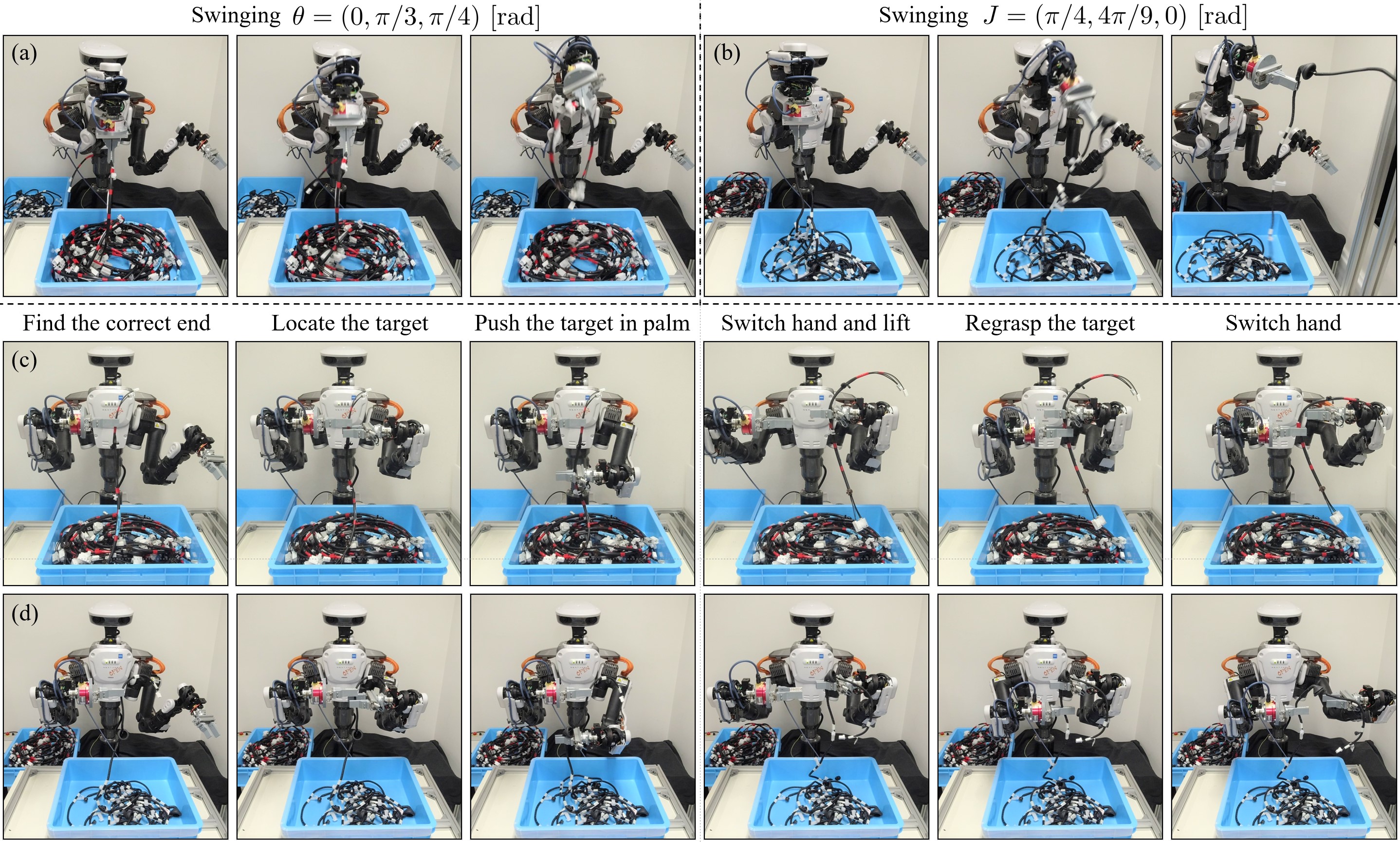}    
    \caption{\textbf{Disentangling motion primitives.} (a-b) Swing motions using different parameters for two types of wire harnesses. The robot's movements can rapidly separate the target from entanglement. (c-d) Regrasping motions for two types of wire harnesses.}
    \label{fig:motion} 
\end{figure*}

\subsection{Dynamic Manipulation for Deformable Objects}
    Deformable object manipulation has primarily focused on 1D objects such as cables and ropes \cite{lui2013tangled,grannen2020untangling,lim2022real2sim2real,she2021cable,ma2023robotic,chi2022irp} and 2D objects like fabrics and cloth \cite{yamakawa2008knotting,yamakawa2011dynamic,ha2022flingbot,hoque2020visuospatial}. Various studies have successfully accomplished challenging manipulation tasks in these domains. Grannen et al. \cite{grannen2020untangling} proposed a method to untangle knots based on learned keypoints and bimanual manipulation. Seita et al. \cite{seita2020deep} learned a sequential policy that utilizes pick-and-place actions to smooth cloth. However, quasi-static manipulation have difficulties in dealing with heavy self-occlusion in 1D deformable objects and higher dimensions in cloth or fabric. Dynamic manipulation, which involves higher velocities and considers inertia effects, has shown effectiveness in manipulating deformable objects \cite{yamakawa2008knotting,chi2022irp,ha2022flingbot,yamakawa2011dynamic,hoque2020visuospatial}. Chi et al. \cite{chi2022irp} developed an iterative policy for goal-conditional manipulation using visual feedback. Chen et al. \cite{chen2022efficiently} proposed a learning framework that enables a single arm to dynamically smooth cloth. Yamakawa et al. \cite{yamakawa2008knotting,yamakawa2011dynamic} introduced an analytic control algorithm for performing high-speed manipulation tasks. Viswanath et al. \cite{viswanath2022autonomously} proposed a shaking motion to dynamically reduce loops and reveal knots in entangled cables. Building upon the advantages of dynamic manipulation, we craft effective motion primitives incorporating high speed to effectively disentangled the wire harnesses. 
    
\subsection{Manipulation of Multiple Deformable Objects}

    Manipulating multiple deformable objects poses unique challenges. The solutions must simultaneously consider the heacy occlusion and rich contact formed by a large number of objects and the entanglement issues. Viswanath et al. \cite{viswanath2021disentangling} proposed to disentangle multi-cable knots by task-relevant keypoint prediction and knot graph representation. Huang et al. \cite{huang2023untangling} presented a method for untangling multiple deformable linear objects by tracing the topological representation. Other studies focus on scenes with a higher degree of entanglmenet in food industry. Ray et al. \cite{ray2020robotic} proposed a method using a designed two-finger gripper to untangle herbs from a pile. Takahashi et al. \cite{takahashi2021target} developed a learning-based separation strategy for grasping a specified mass of small food pieces. Although studies has addressed factory automation problems related to wire harnesses \cite{guo2022visual,jiang2011robotized,zhou2020practical}, robotic bin picking using entangled wire harness remains relatively unexplored. In this paper, we propose a novel and efficient bin picking strategy specifically designed for wire harnesses and improve the performance compared with our previous work \cite{zhang2022learning}.

\subsection{Multiple Sensing in Bin Picking}

    Robotic bin picking relies on computer vision processing for object recognition and grasp planning. However, the integration of other sensory inputs can enhance the robustness for handling heavy occlusion and the intricate objects properties. Haptic feedback is widely adopted in industrial robots for failure detection. Moreira et al. \cite{moreira2016assessment} utilized a force sensor to assess the success of picking operations. Hegemann et al. \cite{hegemann2022learning} proposed a failure detection algorithm for grasping based on both visual and haptic inputs. Studies also use tactile information to guides the bin picking process instead of vision. Ishige et al. \cite{ishige2020blind} proposed a vision-less system that relies solely on tactile feedback to pick small bulked objects. The combination of vision and haptic signals in a closed-loop manner shows great promise in the field of smart manufacturing. By leveraging these multiple modalities, industrial bin picking systems can achieve higher robustness and precision. In this paper, we leverage force signals to complement the vision module in handling challenging cases of picking entangled wire harnesses.

\section{Method}

    The goal of this study is to grasp wire harnesses individually from dense clutter. In this section, we present the manipulation planning of two disentangling motion primitives, the closed-loop workflow with force monitoring and online parameter tuning process. These modules collaborate together to ensure the robustness, effectiveness and generalization of wire harnesses picking.

\subsection{Dynamic Manipulation for Disentangling}

    We design two motion primitives. \textbf{Swing}, which involves high speed and acceleration, can effectively separate the entangled wire harnesses. \textbf{Regrasping} motion enables the robot to adjust the grasp pose from the end of the objects to the middle, making it effective for subsequent actions. 
    
\subsubsection{Swing}
    We use a parametric action primitive to describe the movement of the robot. The action space for the swing primitive is $a=(\theta,\omega,n)$, where $\theta=(\theta_3,\theta_4,\theta_5)$ is the moving angles for the $i$-th joint in one robot arm. $\omega$ denotes the permissible angular velocity across all joints while $n$ indicates the number of times the swing motion is repeated. Specifically, $\theta_5$ denotes the angle for the yaw rotation for the last joint of the robot arm, which can be seen as a ``spinning'' motion. Meanwhile, $\theta_4,\theta_5$ are roll and pitch angles for the last arm joint and can perform a ``whipping'' motion. Three joints of the robot arm moves simultaneously and dynamically extract the objects from the clutter. Note that we initially set the values of $\theta$ and they can be tuned during the execution. The swing motion is illustrated in Fig. \ref{fig:motion}(a-b). 

\subsubsection{Regrasp}
    Regrasp can switch the grasp pose to the middle of the target after the robot grasps the end of the object. Regrasping relies on force feedback rather than vision. The process is illustrated in Fig. \ref{fig:motion}(c-d). Let the right arm of the robot, equipped with a force sensor, acts as the main arm while the left arm is the support arm. The main arm first grasps a wire harness and move to a pre-determine pose. Next, to determine the correct end of the object, the main arm's wrist spins by $\pi$ [rad] and we record the torque signals of both poses. The correct object end is determined by the minimal torque. Then, the support arm move to the pose where its gripper is below the main arm's gripper, moves downward and ensures the object is securely held in the gripper. It then closes the gripper and pulls the object upward. Finally, the main arm moves to the pose where its gripper is below the support arm's gripper and grasp the objects. After the support arm returns to its the initial pose, the regrasping attempt is completed and the main arm successfully adjusts the grasp pose.

\begin{algorithm}[t]
    \SetKwInput{KwInput}{input} 
    \SetKwInput{KwOutput}{output} 
    \SetKwRepeat{Do}{do}{while} 
    \SetKwFunction{FGE}{FGE} 
    
    \KwInput{Depth map, $F_\text{stop}$, $F_\text{fail}$}
    Detect grasp pose from input depth map\;
    $N_\text{transport} \gets 0, L \gets$ empty list\;
    \While{True}
    {
        \texttt{Lift} with force monitoring\;
        \uIf{\rm $F_z^0,F_z^1,...,F_z^t<F_\text{stop}$}
        {
            {Stop}\;
            \texttt{Swing}($\theta,\omega,n$)\;
        }
        \uElseIf{$\dot{F_z}\rightarrow 0$ \rm or $N_\text{transport}>2$ }
        {
            \texttt{Regrasp}\;
        }
        \texttt{Swing}($(0,0,\hat{\theta_5}),\hat{\omega},2$)\;
        \texttt{Transport} with force monitoring\;
        $N_\text{transport} \gets N_\text{transport}+1$\;
        \uIf{\rm $F_z^0,F_z^1,...,F_z^t<F_\text{stop}$ \textbf{or} $F_z^t<F_\text{fail}$}
            {
                Finish\;
            }
        
        $L.$append($F_Z^t$)\;
        \texttt{Update}($F_\text{stop},F_\text{fail},L$)\;
        $\theta \gets \theta+\delta \theta$\;
    }
\caption{Workflow of A Picking Attempt}
\label{alg:system}
\end{algorithm}

\subsection{Closed-Loop System Workflow}

    The workflow of our proposed system is shown in Fig. \ref{fig:overview} and Algorithm \ref{alg:system}. First, we obtain the depth image of the bin filled with wire harnesses and detect a set of collision-free grasp \cite{domae2014fast,zhang2022learning}. The robot then grasps the target and lifts it while monitoring the force $F_z$ in $z$ axis vertically to the workspace. If $F_z$ exceeds the threshold $F_\text{stop}$, the robot immediately stops and performs the swing motion to disentangle the target. Otherwise, If the haptic feedback does not provide an stopping point, meaning either the robot grasps a single wire harness or the regrasping motion is needed. Thus, to determine if the robot should execute the regrasping motion, we use $F_z=\{F_z^0,F_z^1,...,F_z^t\}$ over a time series $t$ recorded during the lifting process. We apply a median filter to $F_z$ and calculate the gradient $\dot{F}_z$. If $\dot{F}_z$ approximates zero, indicating that the target is too long to exert any forces on the gripper, the robot leverages regrasping to change the grasp position to the middle. Then, the robot tries to transport the wire harness to the goal bin while monitoring the force. However, if $F_z$ does not exceed $F_\text{stop}$ during transporting and also does not exceed $F_\text{fail}$ before dropping into the goal bin, the robot performs a successful attempt of picking a single wire harness. Otherwise, we increase the swing parameters $\theta$, adjust the force threshold $F_\text{stop},F_\text{fail}$ and restart from the beginning (lifting while monitoring $F_z$). Additionally, if the robot fails to disentangle the objects after two transporting attempts, it executes the regrasping motion. 
    

\begin{algorithm}[t]
    \SetKwInput{KwInput}{input} 
    \SetKwInput{KwOutput}{output} 
    \SetKwRepeat{Do}{do}{while} 
    \SetKwFunction{FMain}{Update}
    \SetKwProg{Fn}{Function}{:}{}
    
    \Fn{\FMain{\rm $F_\text{stop},F_\text{fail},L$}}
    {
        \uIf{\rm not {stop when \texttt{lift} and \texttt{transport}}}
        {
            \tcp{Minimizes the gradient of $L$}
            $F_\text{fail} \gets \arg\min_{F} \dot{L}$ \;
        }
        \uIf{\rm not stop when \texttt{lift} \textbf{and} stop when \texttt{transport}}
        {
            $F_\text{stop} \leftarrow F_\text{stop} - \delta F$\;
        }
    }
\caption{Online Parameter Tuning}
\label{alg:param}
\end{algorithm}

\subsection{Online Parameter Tuning}

    In Line 15-17 of Algorithm \ref{alg:system}, we introduce an online parameter adjustment algorithm to improve the robustness of the robot. The initial force thresholds are manually set: $F_\text{stop}$ represents the minimal force where the entanglement occurs, while $F_\text{fail}$ approximates the weight of grasping a single object. Algorithm \ref{alg:param} outlines our online parameter tuning process.

    First, before the robot drops objects into the goal bin, we monitor the force $F_z^t$. If the robot successfully transports only one object without any stops during both lifting and transportation ($F_z^t<F_\text{fail}$), we adjust the value of $F_\text{fail}$. After each attempt in this scenario, we obtain a list $L$ of $F_z^t$. $F_\text{fail}$ is updated by minimizing the gradient of $L$ towards zero. The value of $F_\text{fail}$ gradually converges to a value and the updating process is stopped when the gradient no longer changes. Next, if the robot does not stop during lifting ($F_z^t<F_\text{fail}$) but encounters a stop during transporting ($F_z^t<F_\text{stop}$), it indicates that the threshold for detecting entanglement is not sensitive enough and should be tuned to a lower value. In this case, we set a residual force parameter $\delta F$ and adjust the threshold as follows: $F_\text{stop}=F_\text{stop}-\delta F$.

    In addition to the force thresholds, our algorithm also adjusts the swing parameters during each attempt. When the transporting process is unsuccessful, the robot will disentangle the grasped objects using increased swing angles. We increase the $\theta$ by a predefined residual angle $\delta \theta$, while ensuring that the adjustments remain within the velocity limits of the robot's arm. We also implement an additional motion \texttt{Swing}($(0,0,\hat{\theta_5}),\hat{\omega},2$) before the transporting process. This two-way spinning motion with pre-defined $\hat{\theta_5},\hat{\omega}$ can can provide additional assurance.






\begin{figure}[t] 
    \centering
    \includegraphics[width=\linewidth]{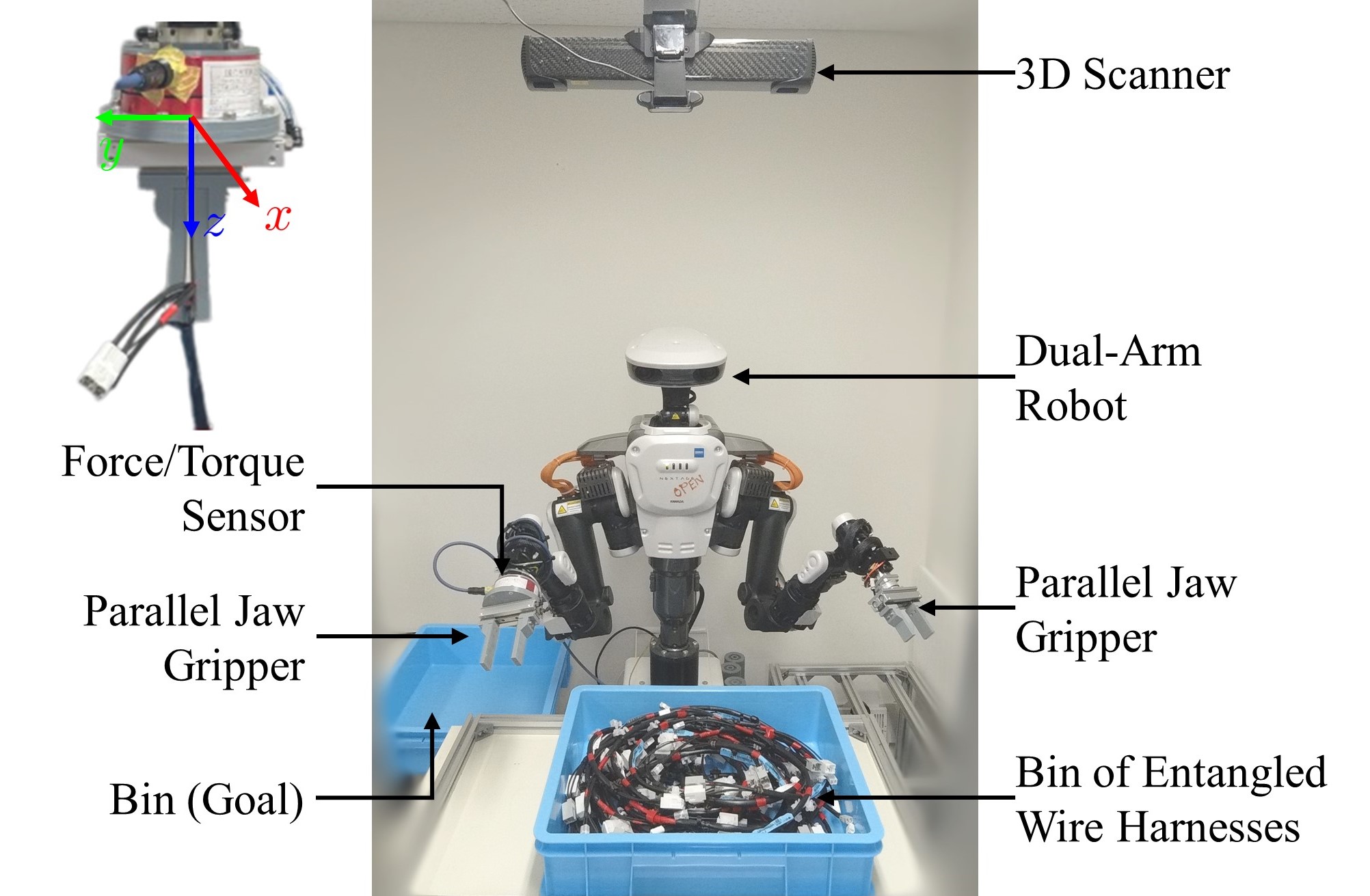}    
    \caption{\small Our experimental setup. }
    \label{fig:setup} 
\end{figure}

\section{Experiments and Results}\label{sec:exp}

\subsection{Experiment Setup}
\begin{figure}[t] 
    \centering`
    \includegraphics[width=\linewidth]{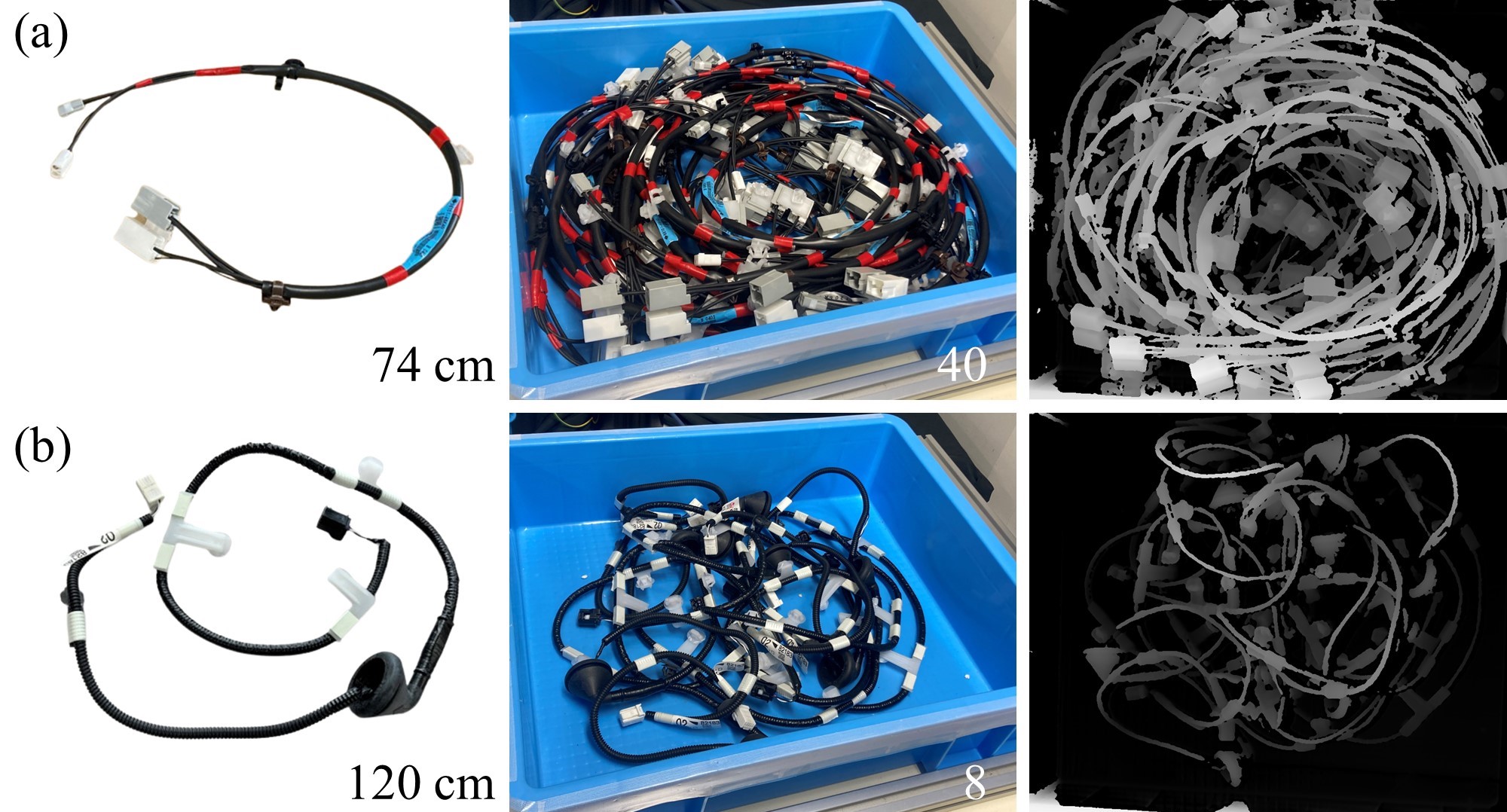}    
    \caption{\textbf{Wire harnesses used in the experiments.} (a) A wire harness that also used to train ASPNet in \cite{zhang2022learning}. (b) A challenging wire harness. }
    \label{fig:obj} 
\end{figure}

    We conduct real-world experiments using two types of wire harnesses measuring 74 cm and 120 cm in length, shown in Fig. \ref{fig:obj}. The bin used in the experiments is filled with a maximum of 8 and 40 objects of each one. We design two specific bin picking tasks for evaluation: 
\begin{itemize}
    \item \textbf{Emptying}: The goal is to completely empty the bin filled with entangled wire harnesses. 
    \item \textbf{Standard}: After each successful picking, we reload the bin with the same number of wire harnesses and randomly shuffle them. This ensures that the robot encounters different entanglement patterns throughout the task.
\end{itemize}

    Our experimental setup is shown in Fig. \ref{fig:setup}. We use a NEXTAGE robot from Kawada Industries Inc. The robot's arms operate within a workspace that is captured as a top-down depth image using a Photoneo PhoXi 3D scanner M. Each arm is equipped with a parallel jaw gripper at its tip. A force sensor DynPick WEF-6A200-4-RCD is mounted at the wrist of the robot's right arm. We use a PC equipped with an Intel Core i7 CPU, 16GB of memory and an Nvidia GeForce 1080 GPU for the physical experiments. We fix the parameters empirically for the experiments. The incremental angles for the swing motion is fixed at $\delta \theta=\pi/18$ [rad], while the incremental forces for online parameter tuning are set to $\delta F = 0.1$ [N]. The initial parameters for swing motion is $\theta_3,\theta_4\,\theta_5=\pi/4,\pi/3,\pi/3$ [rad], $\omega=\pi/2$ [rad/s], $n=2$. The initial force thresholds are $F_\text{stop}=3$ [N], $F_\text{fail}=1$ [N].

    We present two baselines and two ablated version of our method: 

\begin{itemize}
    \item \textbf{Lift-G}: This open-loop method uses Fast Graspability Evaluation (FGE) \cite{domae2014fast} to detect collision-free grasps and directly lifts the target after grasping, without incorporating haptic feedback.
    \item \textbf{Circle-A}: This open-loop method, described in \cite{zhang2022learning}, leverages ASPNet to infer the lowest action complexity of each grasp and execute a circling motion to disentangle the wire harnesses. 
    \item \textbf{Ours-G}: Our closed-loop policy incorporates dynamic and bimanual manipulation with haptic feedback for real-time adjustments. It utilizes the FGE algorithm \cite{domae2014fast} for grasp detection.
    \item \textbf{Ours-A}: Our closed-loop policy optimizes the grasp pose using ASPNet \cite{zhang2022learning}. Instead of simply selecting the top rank from FGE, ASPNet evaluates the action complexity of each grasp and selects the lowest one.
\end{itemize}

\begin{table*}[t]
\renewcommand\arraystretch{1}
\small
    \caption{Success Rate Comparison}
    \centering
    \begin{tabular}{cclcccccc}
        \toprule
        \multirow{2.5}{*}{Object} & \multirow{2.5}{*}{Task} &\multirow{2.5}{*}{Method} & \multirow{2.5}{*}{\# Objects} & \multirow{2.5}{*}{Success Rate (\%)} & \multicolumn{4}{c}{\# Attempts}\\
        \cmidrule(lr){6-9}
        &&&&&Lift &Circle & Swing & Regrasp\\
        \midrule
        \multirow{10}{*}{\includegraphics[width=2cm]{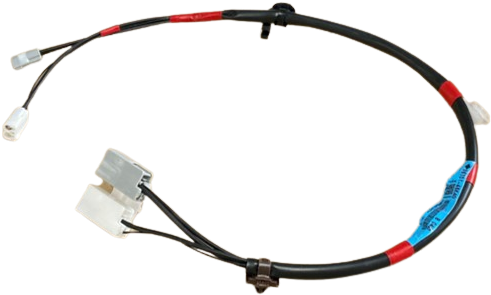}}&\multirow{4}{*}{Emptying} 
        &Lift-G & 15 & 53.6\% (15/28) & 15 & -&-&-\\
        &&Circle-A & 15 & 83.3\% (15/18)& 1 & 14 & - & - \\
        &&\textbf{Ours-G} & 40 & \textbf{94.7\% (36/38)} & 10 & - & 29 & 7 \\
        &&\textbf{Ours-A} & 40 & \textbf{97.4\% (38/39)} & 17 & - & 22 & 5 \\
        \cmidrule(lr){2-9} 
        &\multirow{4}{*}{Standard} 
        &Lift-G & 25 & 23.3\% (7/30) & 7 & -&-&-\\
        &&Circle-A & 25 & 73.3\% (22/30)& - & 22 & - & - \\
        &&Ours-G & 40 & 83.3\% (25/30) & 2 & - & 28 & 5 \\
        &&\textbf{Ours-A} & 40 & \textbf{86.7\% (26/30)} & 5 & - & 19 & 9 \\
        \midrule 
        \multirow{4}{*}{\includegraphics[width=2cm]{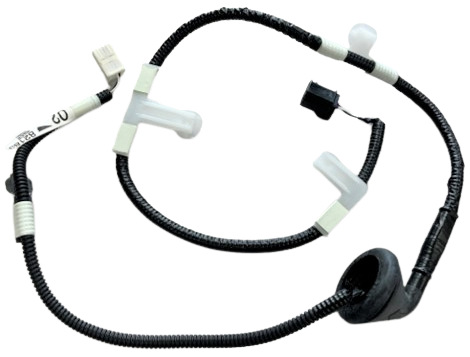}}& \multirow{2}{*}{Emptying} 
        &Circle-A & 8 & 0.0\% (0/20)&  - & - & - & - \\
        &&\textbf{Ours-A} & 8 & \textbf{80.0\% (16/20)} & - & -&5&12\\
        \cmidrule(lr){2-9}
        &\multirow{2}{*}{Standard}
        &Circle-A & 8 & 0.0\% (0/20)&  - & - & - & - \\
        &&\textbf{Ours-A} & 8 & \textbf{65.0\% (13/20)} & - & -&9&10\\
        \bottomrule
    \end{tabular}
    \label{tab:result}
\end{table*}


\subsection{Comparisons with Baselines}

    Table \ref{tab:result} presents the performance comparison among our methods and the baselines. In the emptying task, both Ours-G and Ours-A demonstrate significant improvements in success rates compared to the baselines. The average success rates of emptying task achieve 96.1\% and 80\% respectvely for two types of wire harnesses. Since Lift-G and Circle-A without haptic feedback are unable to handle dense clutter, we evaluate their performance using less objects instead. Notably, our policy outperforms the other methods, even under a higher degree of entanglement due to effective disentangling motion primitives. Especially, for wire harnesses with a length of 120 cm, our policy improves the success rate from 0\% to 80\%. The swing motion plays a crucial role in separating the objects, and the regrasping motion leads to a remarkable success rate increase for longer wire harnesses. 
    
\begin{figure*}[t] 
    \centering
    \includegraphics[width=\linewidth]{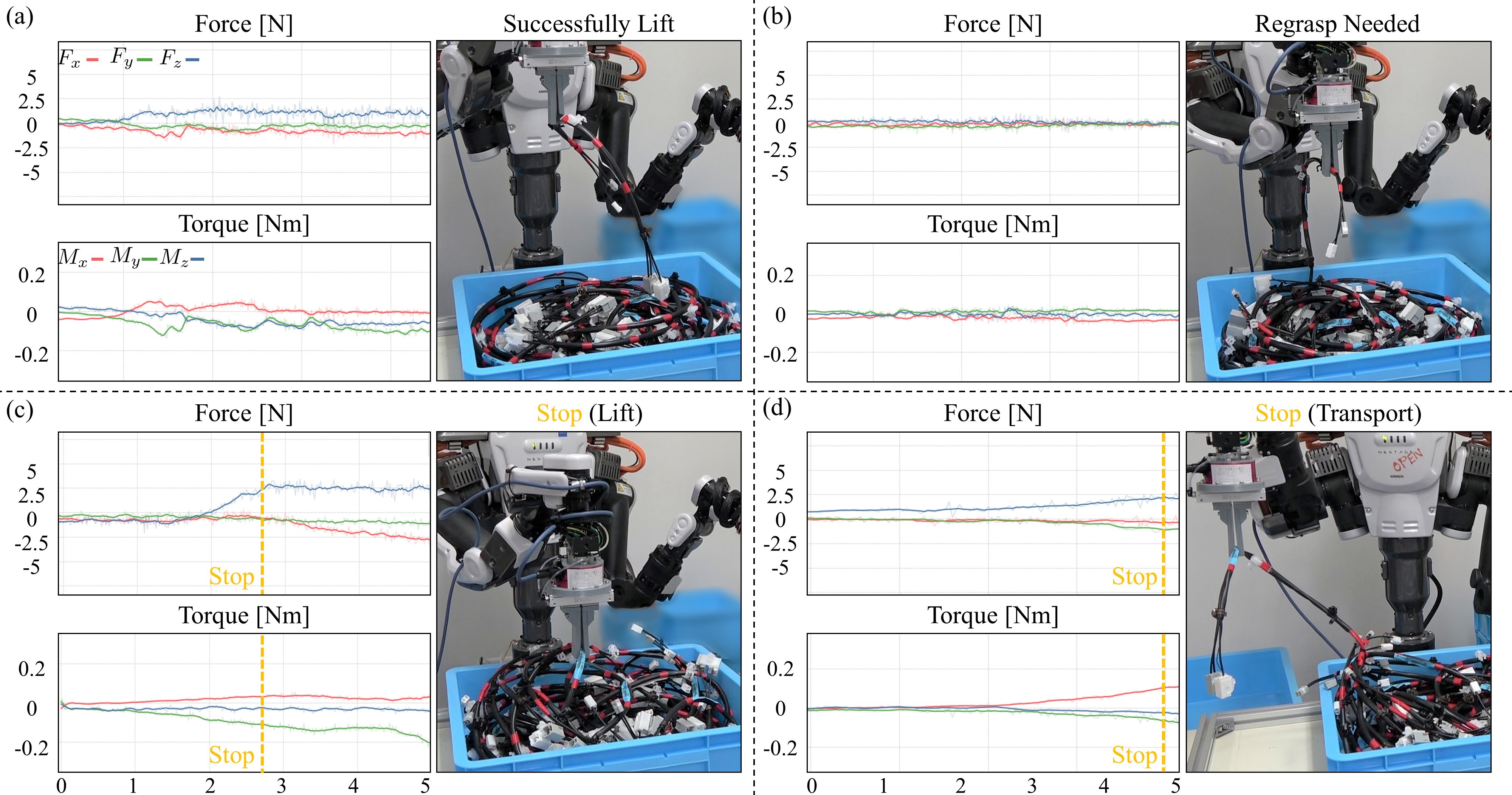}    
    \caption{\textbf{Visualized outputs from the force sensor during different scenarios.} (a) The robot successfully grasps and lifts an isolated object without entanglement, as indicated by the smooth increase in $F_z$ (blue line). (b) When grasping at the end of an object, $F_z$ remains near zero without a significant increase. (c) The robot detects entanglement (marked in yellow) while lifting the target and immediately stops, as $F_z$ shows a sharp increase exceeding the threshold $F_\textbf{stop}$. (d) During transportation of the target to the goal bin, the robot stops after detecting entanglement, again indicated by $F_z$ exceeding the threshold $F_\textbf{stop}$.}
    \label{fig:ret} 
\end{figure*}

    For the standard task where the robot must confront more complex entanglement patterns, our policy demonstrates a significant improvement in success rates for both types of objects. Different from the emptying task, where fewer objects remain in the bin at the later half of the task, the standard task keeps a higher degree of entanglement throughout the picking process. The real-time haptic feedback mechanism facilitates the recovery from failed disentangling actions. Every module in our proposed closed-loop system works collaboratively to achieve efficient bin picking from perception to manipulation planning. Additionally, Ours-A outperforms Ours-G in success rates overall since Ours-A can avoid grasping the ends of the objects, leading to more sufficient disentangling actions.

\subsection{Benefits of Closed-Loop system with Haptic Feedback}

    We also visualize the force monitoring process during execution. In Fig. \ref{fig:ret}, we illustrate the robot's actions and the corresponding force readings. The robot utilizes force signals to detect whether the entanglement occurs (Fig. \ref{fig:ret}(c-d) or not (Fig. \ref{fig:ret}(a-b)). By incorporating force feedback, we effectively mitigate errors and mistakes that may arise when relying solely on visual predictions. 

\begin{figure}[H] 
    \centering
    \includegraphics[width=\linewidth]{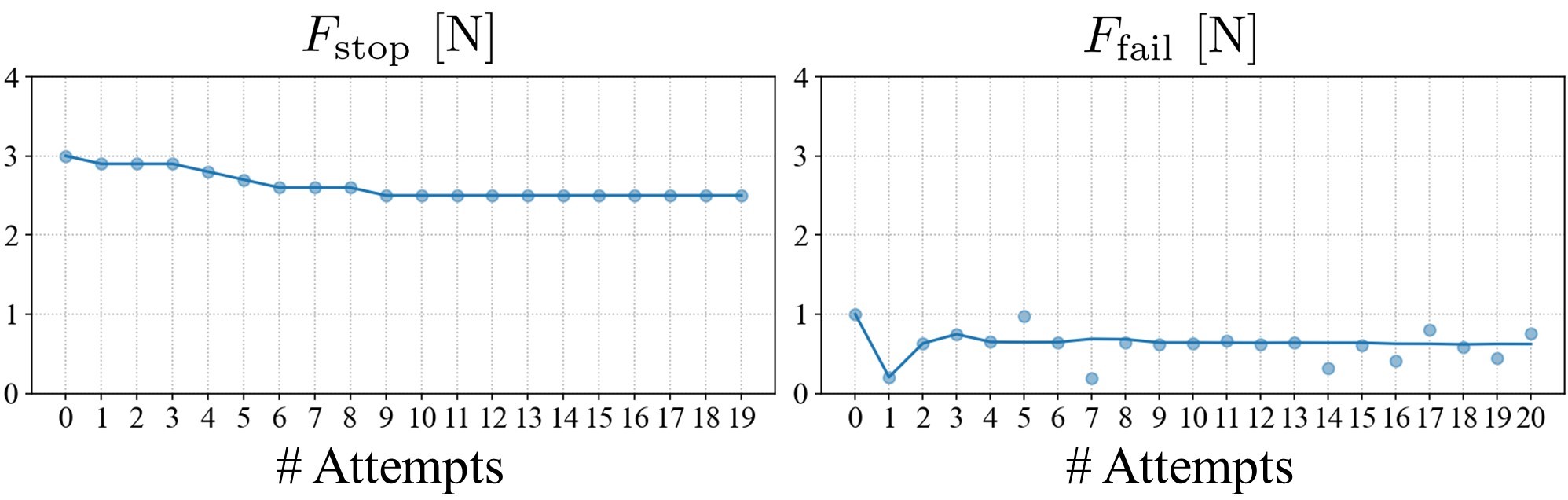}    
    \caption{Results of the online tuning procedure. }
    \label{fig:tuning} 
\end{figure}

    Fig. \ref{fig:tuning} provides the adjustment of force thresholds throughout the consecutive picking process. The force threshold $F_\text{stop}$ gradually decreases over time and eventually stabilizes at a certain value. This value represents the minimum force at which entanglement occurs. On the other hand, the distribution of $F_\text{fail}$ is more scattered, and we optimize it by minimizing the gradient to zero. The optimization process leads to the convergence of the threshold to a stable value, which closely approximates the weight of a single ob. The online parameter tuning acts as a valuable supervisor, enhancing the overall performance and generalization of our system when adapting to previously unseen objects.
    
\subsection{Benefits of Dynamic Motion Primitives}
    
    Table \ref{tab:result} also includes the attempt numbers for performing each disentangling motion primitive in the baselines and our methods. The results demonstrate that incorporating swing and regrasping achieves higher success rates compared to the circling motion. The swing motion effectively disentangles the target and loosens dense entanglements, particularly for longer wire harnesses. Additionally, regrasping enhances the accuracy of the swing motion by switching to a more suitable grasping position on the target. We can also observe that the longer wire harness has more regrasping attempts. This dual-arm manipulation can effectively addresses the issue of length and also allows for directly pulling the target from the entanglement. 

    In addition, we evaluate the force applied to the objects during the circling motion and our proposed dynamic motion primitives. The force readings show that the proposed motion primitives can successfully complete a picking attempt with a force of only 5 [N], which almost the same as the quasi-static circling motion. Applying less force to the objects reduces the potential damage to the wire harnesses, thereby minimizing wear and tear during the assembly process.

\begin{figure}[H] 
    \centering
    \includegraphics[width=\linewidth]{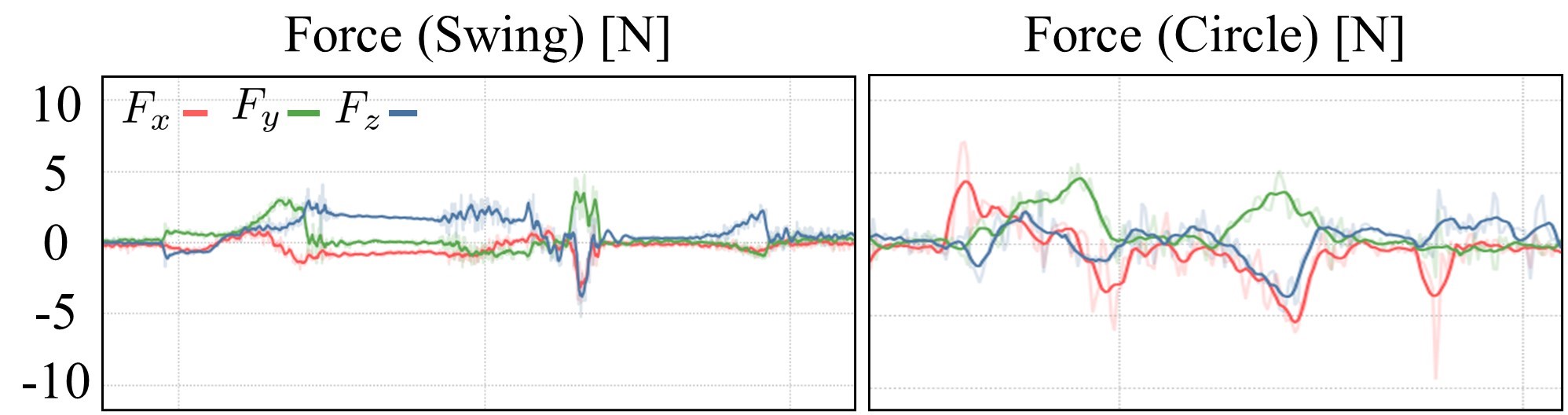}    
    \caption{\textbf{Force comparison between swing and circling motions.} Swing motion exerts a moderate force on the wire harnesses, thereby preventing damage to them during the picking process.}
    \label{fig:act-force} 
\end{figure}

\subsection{Benefits of ASPNet}

    Table \ref{tab:aspnet-ac} shows the normalized action complexity predicted by ASPNet \cite{zhang2022learning}. ASPNet effectively predicts that longer objects require more complex actions. The result demonstrates that ASPNet can effectively predict the complexity of the entanglment from observation. Although we did not specifically match the action complexity with each action in this study, we leverage this learned vision model to assist in choosing more suitable grasps. In table \ref{tab:result}, Ours-A completes the task with more lifting attempts and fewer swing attempts than Ours-G. This demonstrates ASPNet can seek objects of a lower level of the entanglement, making the picking efficiency higher than using FGE. 
    
\begin{table}[H]
\renewcommand\arraystretch{1}
\footnotesize
    \caption{Normalized Action Complexity Predicted by ASPNet}
    \centering
    \begin{tabular}{ccc}
        \toprule
        \multirow{6}{*}{\# Objects} & \multicolumn{2}{c}{Action Complexity}\\
        \cmidrule(lr){2-3}
        & \includegraphics[width=2cm]{fig/obj_medium.png} & \includegraphics[width=2cm]{fig/obj_long.png} \\
        \midrule 
        5 & 0.133 & 0.800\\
        10 & 0.467 & 0.767\\
        15 & 0.483 & 0.800\\
        \bottomrule
    \end{tabular}
    \label{tab:aspnet-ac}
\end{table}

\begin{figure}[H] 
    \centering
    \includegraphics[width=\linewidth]{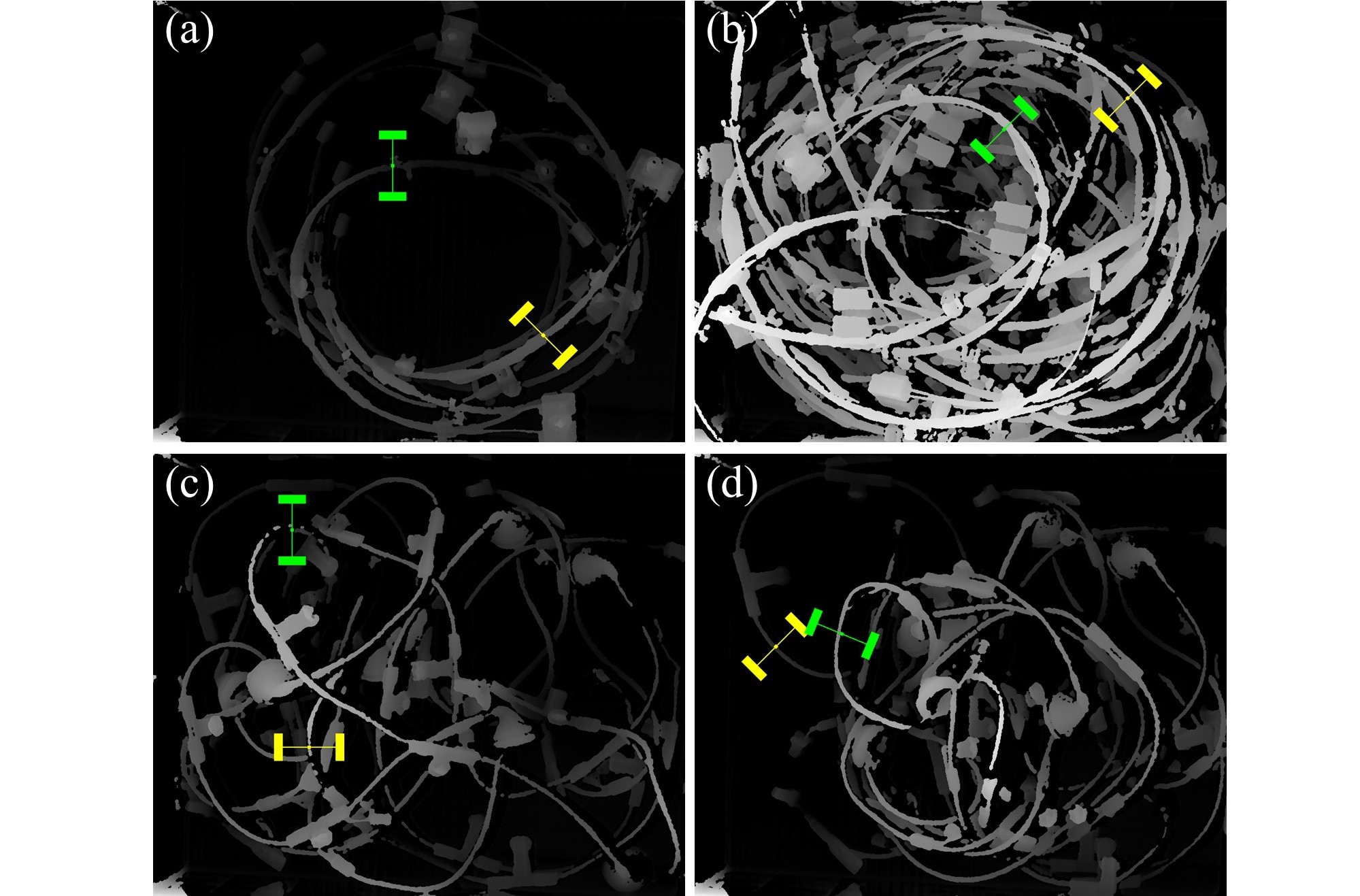}    
    \caption{\textbf{Grasps computed from FGE (yellow) and ASPNet (green).} ASPNet tends to find objects located at the top of the heap and aims to grasp them at their middle part.}
    \label{fig:grasp-cprs} 
\end{figure}

    Figure \ref{fig:grasp-cprs} illustrates the grasp poses detected from the same depth image using the FGE (-G) \cite{domae2014fast} and ASPNet (-A) \cite{zhang2022learning}. Grasp poses marked in green are detected using ASPNet, which always find the objects located at the top of the clutter and grasps them at the middle. On the other hand, FGE detects grasp poses marked in yellow, which have a high score for avoiding collisions with the gripper but does not consider the entanglement issue, resulting in lower picking efficiency.
 
\section{Failure Modes and Discussion}

    Table \ref{tab:failure} presents the four failure modes and their corresponding frequencies when using our methods, Ours-G and Our-A.

\begin{enumerate}[(A)]
    \item The robot transports nothing to the goal bin due to \textbf{grasp failure}. 
    \item The robot transports nothing to the goal bin due to \textbf{swing failure}. Swing motion sometimes makes the other objects sprung out of the bin. Additionally, there are cases where the objects slipped from the gripper during high-speed swing motions.
    \item The robot transports nothing to the goal bin due to \textbf{regrasping failure}. After the main arm of the robot moves to the initial pose, in cases where the target is not aligned vertically with the workspace, the support arm cannot accurately locate the pose of the target.
    \item The robot transports multiple objects into the goal bin due to \textbf{recovery error}. Force monitoring fails to detect the entanglement. 
\end{enumerate}
    
\begin{table}[H]
\renewcommand\arraystretch{1}
\small
    \caption{Failure Cases in Ours-G/Ours-A and Their Frequencies}
    \centering
    \begin{tabular}{lcc}
        \toprule
        \multirow{6}{*}{Failure Mode} & \multicolumn{2}{c}{Frequency}\\
        \cmidrule(lr){2-3}
        & \includegraphics[width=2cm]{fig/obj_medium.png} & \includegraphics[width=2cm]{fig/obj_long.png} \\
        \midrule 
        (A) Grasp failure & 2/127 & 1/40\\
        (B) Swing Failure & 1/127 & 2/40 \\
        (C) Regrasping failure & 4/127 & 4/40 \\
        (D) Recovery Failure & 5/127 & 4/40 \\
        \bottomrule
    \end{tabular}
    \label{tab:failure}
\end{table}

    Table \ref{tab:failure} shows the frequency of each failure case. For long objects, the occurrence of regrasping failure and recovery failure is significantly higher compared to another type. It suggests that achieving robust and successful regrasping solely relying on force feedback without visual feedback is challenging. 




\section{Conclusion}\label{sec:con}
    
    In this paper, we present a novel bin picking system specifically for grasping and separating entangled wire harnesses. Our closed-loop system utilizes dynamic manipulation with haptic feedback, enabling successful handling of complex entanglement scenarios. Through real-world experiments, we demonstrate the effectiveness of our policy in disentangling various wire harnesses with high success rates. In future work, we will address failure cases by incorporating vision-guided regrasping motion or vision-haptic fusion policies. Additionally, we will enhance the perception module to obtain more precise and interpretable visual representations of the entangled deformable objects. Moreover, we will focus on optimizing the parameters of dynamic motion primitives to ensure both accuracy and safety in bin picking.

\section*{Acknowledgement}
    This research was supported by Toyota Motor Corporation.    
    
\bibliographystyle{ieeetr}
\bibliography{ebibsample.bib}

\begin{thebibliography}{10}

\bibitem{kirkegaard2006bin}
J.~Kirkegaard and T.~B. Moeslund, ``Bin-picking based on harmonic shape
  contexts and graph-based matching,'' in {\em 18th International Conference on
  Pattern Recognition (ICPR'06)}, vol.~2, pp.~581--584, IEEE, 2006.

\bibitem{liu2012fast}
M.-Y. Liu, O.~Tuzel, A.~Veeraraghavan, Y.~Taguchi, T.~K. Marks, and
  R.~Chellappa, ``Fast object localization and pose estimation in heavy clutter
  for robotic bin picking,'' {\em The International Journal of Robotics
  Research}, vol.~31, no.~8, pp.~951--973, 2012.

\bibitem{buchholz2013efficient}
D.~Buchholz, M.~Futterlieb, S.~Winkelbach, and F.~M. Wahl, ``Efficient
  bin-picking and grasp planning based on depth data,'' in {\em 2013 IEEE
  International Conference on Robotics and Automation}, pp.~3245--3250, IEEE,
  2013.

\bibitem{domae2014fast}
Y.~Domae, H.~Okuda, Y.~Taguchi, K.~Sumi, and T.~Hirai, ``Fast graspability
  evaluation on single depth maps for bin picking with general grippers,'' in
  {\em 2014 IEEE International Conference on Robotics and Automation (ICRA)},
  pp.~1997--2004, IEEE, 2014.

\bibitem{harada2016initial}
K.~Harada, W.~Wan, T.~Tsuji, K.~Kikuchi, K.~Nagata, and H.~Onda, ``Initial
  experiments on learning-based randomized bin-picking allowing finger contact
  with neighboring objects,'' in {\em 2016 IEEE International Conference on
  Automation Science and Engineering (CASE)}, pp.~1196--1202, IEEE, 2016.

\bibitem{matsumura2018learning}
R.~Matsumura, K.~Harada, Y.~Domae, and W.~Wan, ``Learning based industrial
  bin-picking trained with approximate physics simulator,'' in {\em
  International Conference on Intelligent Autonomous Systems}, pp.~786--798,
  Springer, 2018.

\bibitem{tachikake2020learning}
H.~Tachikake and W.~Watanabe, ``A learning-based robotic bin-picking with
  flexibly customizable grasping conditions,'' in {\em 2020 IEEE/RSJ
  International Conference on Intelligent Robots and Systems (IROS)},
  pp.~9040--9047, IEEE, 2020.

\bibitem{zhang2022learning}
X.~Zhang, Y.~Domae, W.~Wan, and K.~Harada, ``Learning efficient policies for
  picking entangled wire harnesses: An approach to industrial bin picking,''
  {\em IEEE Robotics and Automation Letters}, vol.~8, no.~1, pp.~73--80, 2022.

\bibitem{choi2012voting}
C.~Choi, Y.~Taguchi, O.~Tuzel, M.-Y. Liu, and S.~Ramalingam, ``Voting-based
  pose estimation for robotic assembly using a 3d sensor,'' in {\em 2012 IEEE
  International Conference on Robotics and Automation}, pp.~1724--1731, IEEE,
  2012.

\bibitem{yang2021probabilistic}
J.~Yang, D.~Li, and S.~L. Waslander, ``Probabilistic multi-view fusion of
  active stereo depth maps for robotic bin-picking,'' {\em IEEE Robotics and
  Automation Letters}, vol.~6, no.~3, pp.~4472--4479, 2021.

\bibitem{liu20216d}
D.~Liu, S.~Arai, Y.~Xu, F.~Tokuda, and K.~Kosuge, ``6d pose estimation of
  occlusion-free objects for robotic bin-picking using ppf-meam with 2d images
  (occlusion-free ppf-meam),'' {\em IEEE Access}, vol.~9, pp.~50857--50871,
  2021.

\bibitem{dupuis2008two}
D.~C. Dupuis, S.~L{\'e}onard, M.~A. Baumann, E.~A. Croft, and J.~J. Little,
  ``Two-fingered grasp planning for randomized bin-picking,'' in {\em Proc. of
  the Robotics: Science and Systems 2008 Manipulation Workshop-Intelligence in
  Human Environments}, 2008.

\bibitem{buchholz2014combining}
D.~Buchholz, D.~Kubus, I.~Weidauer, A.~Scholz, and F.~M. Wahl, ``Combining
  visual and inertial features for efficient grasping and bin-picking,'' in
  {\em 2014 IEEE international conference on robotics and automation (ICRA)},
  pp.~875--882, IEEE, 2014.

\bibitem{harada2013probabilistic}
K.~Harada, K.~Nagata, T.~Tsuji, N.~Yamanobe, A.~Nakamura, and Y.~Kawai,
  ``Probabilistic approach for object bin picking approximated by cylinders,''
  in {\em 2013 IEEE International Conference on Robotics and Automation},
  pp.~3742--3747, IEEE, 2013.

\bibitem{mahler2017dex}
J.~Mahler, J.~Liang, S.~Niyaz, M.~Laskey, R.~Doan, X.~Liu, J.~A. Ojea, and
  K.~Goldberg, ``Dex-net 2.0: Deep learning to plan robust grasps with
  synthetic point clouds and analytic grasp metrics,'' in {\em Robotics:
  Science and Systems XIII, Massachusetts Institute of Technology, Cambridge,
  Massachusetts, USA, July 12-16, 2017}, 2017.

\bibitem{tong2021dig}
Z.~Tong, Y.~H. Ng, C.~H. Kim, T.~He, and J.~Seo, ``Dig-grasping via direct
  quasistatic interaction using asymmetric fingers: An approach to effective
  bin picking,'' {\em IEEE Robotics and Automation Letters}, vol.~6, no.~2,
  pp.~3033--3040, 2021.

\bibitem{morino2020sheet}
K.~Morino, S.~Kikuchi, S.~Chikagawa, M.~Izumi, and T.~Watanabe, ``Sheet-based
  gripper featuring passive pull-in functionality for bin picking and for
  picking up thin flexible objects,'' {\em IEEE Robotics and Automation
  Letters}, vol.~5, no.~2, pp.~2007--2014, 2020.

\bibitem{ishige2020blind}
M.~Ishige, T.~Umedachi, Y.~Ijiri, T.~Taniguchi, and Y.~Kawahara, ``Blind bin
  picking of small screws through in-finger manipulation with compliant robotic
  fingers,'' in {\em 2020 IEEE/RSJ International Conference on Intelligent
  Robots and Systems (IROS)}, pp.~9337--9344, IEEE, 2020.

\bibitem{matsumura2019learning}
R.~Matsumura, Y.~Domae, W.~Wan, and K.~Harada, ``Learning based robotic
  bin-picking for potentially tangled objects,'' in {\em 2019 IEEE/RSJ
  International Conference on Intelligent Robots and Systems (IROS)},
  pp.~7990--7997, IEEE, 2019.

\bibitem{zhang2021topological}
X.~Zhang, K.~Koyama, Y.~Domae, W.~Wan, and K.~Harada, ``A topological solution
  of entanglement for complex-shaped parts in robotic bin-picking,'' in {\em
  2021 IEEE 17th International Conference on Automation Science and Engineering
  (CASE)}, pp.~461--467, IEEE, 2021.

\bibitem{zhang2023learning}
X.~Zhang, Y.~Domae, W.~Wan, and K.~Harada, ``Learning to dexterously pick or
  separate tangled-prone objects for industrial bin picking,'' {\em arXiv
  preprint arXiv:2302.08152}, 2023.

\bibitem{guo2022visual}
J.~Guo, J.~Zhang, Y.~Gai, D.~Wu, and K.~Chen, ``Visual recognition method for
  deformable wires in aircrafts assembly based on sequential segmentation and
  probabilisitic estimation,'' in {\em 2022 IEEE 6th Information Technology and
  Mechatronics Engineering Conference (ITOEC)}, vol.~6, pp.~598--603, IEEE,
  2022.

\bibitem{jiang2011robotized}
X.~Jiang, K.-m. Koo, K.~Kikuchi, A.~Konno, and M.~Uchiyama, ``Robotized
  assembly of a wire harness in a car production line,'' {\em Advanced
  Robotics}, vol.~25, no.~3-4, pp.~473--489, 2011.

\bibitem{zhou2020practical}
H.~Zhou, S.~Li, Q.~Lu, and J.~Qian, ``A practical solution to deformable linear
  object manipulation: A case study on cable harness connection,'' in {\em 2020
  5th International Conference on Advanced Robotics and Mechatronics (ICARM)},
  pp.~329--333, IEEE, 2020.

\bibitem{lui2013tangled}
W.~H. Lui and A.~Saxena, ``Tangled: Learning to untangle ropes with rgb-d
  perception,'' in {\em 2013 IEEE/RSJ International Conference on Intelligent
  Robots and Systems}, pp.~837--844, IEEE, 2013.

\bibitem{grannen2020untangling}
J.~Grannen, P.~Sundaresan, B.~Thananjeyan, J.~Ichnowski, A.~Balakrishna,
  M.~Hwang, V.~Viswanath, M.~Laskey, J.~E. Gonzalez, and K.~Goldberg,
  ``Untangling dense knots by learning task-relevant keypoints,'' in {\em
  Proceedings of Robotics: Science and Systems (RSS)}, 2020.

\bibitem{lim2022real2sim2real}
V.~Lim, H.~Huang, L.~Y. Chen, J.~Wang, J.~Ichnowski, D.~Seita, M.~Laskey, and
  K.~Goldberg, ``Real2sim2real: Self-supervised learning of physical
  single-step dynamic actions for planar robot casting,'' in {\em 2022
  International Conference on Robotics and Automation (ICRA)}, pp.~8282--8289,
  IEEE, 2022.

\bibitem{she2021cable}
Y.~She, S.~Wang, S.~Dong, N.~Sunil, A.~Rodriguez, and E.~Adelson, ``Cable
  manipulation with a tactile-reactive gripper,'' {\em The International
  Journal of Robotics Research}, vol.~40, no.~12-14, pp.~1385--1401, 2021.

\bibitem{ma2023robotic}
Z.~Ma and J.~Xiao, ``Robotic perception-motion synergy for novel rope wrapping
  tasks,'' {\em IEEE Robotics and Automation Letters}, 2023.

\bibitem{chi2022irp}
C.~Chi, B.~Burchfiel, E.~Cousineau, S.~Feng, and S.~Song, ``Iterative residual
  policy for goal-conditioned dynamic manipulation of deformable objects,'' in
  {\em Proceedings of Robotics: Science and Systems}, 2022.

\bibitem{yamakawa2008knotting}
Y.~Yamakawa, A.~Namiki, M.~Ishikawa, and M.~Shimojo, ``Knotting manipulation of
  a flexible rope by a multifingered hand system based on skill synthesis,'' in
  {\em 2008 IEEE/RSJ International Conference on Intelligent Robots and
  Systems}, pp.~2691--2696, IEEE, 2008.

\bibitem{yamakawa2011dynamic}
Y.~Yamakawa, A.~Namiki, and M.~Ishikawa, ``Dynamic manipulation of a cloth by
  high-speed robot system using high-speed visual feedback,'' {\em IFAC
  Proceedings Volumes}, vol.~44, no.~1, pp.~8076--8081, 2011.

\bibitem{ha2022flingbot}
H.~Ha and S.~Song, ``Flingbot: The unreasonable effectiveness of dynamic
  manipulation for cloth unfolding,'' in {\em Conference on Robot Learning},
  pp.~24--33, PMLR, 2022.

\bibitem{hoque2020visuospatial}
R.~Hoque, D.~Seita, A.~Balakrishna, A.~Ganapathi, A.~K. Tanwani, N.~Jamali,
  K.~Yamane, S.~Iba, and K.~Goldberg, ``Visuospatial foresight for multi-step,
  multi-task fabric manipulation,'' {\em arXiv preprint arXiv:2003.09044},
  2020.

\bibitem{seita2020deep}
D.~Seita, A.~Ganapathi, R.~Hoque, M.~Hwang, E.~Cen, A.~K. Tanwani,
  A.~Balakrishna, B.~Thananjeyan, J.~Ichnowski, N.~Jamali, {\em et~al.}, ``Deep
  imitation learning of sequential fabric smoothing from an algorithmic
  supervisor,'' in {\em 2020 IEEE/RSJ International Conference on Intelligent
  Robots and Systems (IROS)}, pp.~9651--9658, IEEE, 2020.

\bibitem{chen2022efficiently}
L.~Y. Chen, H.~Huang, E.~Novoseller, D.~Seita, J.~Ichnowski, M.~Laskey,
  R.~Cheng, T.~Kollar, and K.~Goldberg, ``Efficiently learning single-arm fling
  motions to smooth garments,'' {\em arXiv preprint arXiv:2206.08921}, 2022.

\bibitem{viswanath2022autonomously}
V.~Viswanath, K.~Shivakumar, J.~Kerr, B.~Thananjeyan, E.~Novoseller,
  J.~Ichnowski, A.~Escontrela, M.~Laskey, J.~E. Gonzalez, and K.~Goldberg,
  ``Autonomously untangling long cables,'' {\em arXiv preprint
  arXiv:2207.07813}, 2022.

\bibitem{viswanath2021disentangling}
V.~Viswanath, J.~Grannen, P.~Sundaresan, B.~Thananjeyan, A.~Balakrishna,
  E.~Novoseller, J.~Ichnowski, M.~Laskey, J.~E. Gonzalez, and K.~Goldberg,
  ``Disentangling dense multi-cable knots,'' in {\em 2021 IEEE/RSJ
  International Conference on Intelligent Robots and Systems (IROS)},
  pp.~3731--3738, IEEE, 2021.

\bibitem{huang2023untangling}
X.~Huang, D.~Chen, Y.~Guo, X.~Jiang, and Y.~Liu, ``Untangling multiple
  deformable linear objects in unknown quantities with complex backgrounds,''
  {\em IEEE Transactions on Automation Science and Engineering}, 2023.

\bibitem{ray2020robotic}
P.~Ray and M.~J. Howard, ``Robotic untangling of herbs and salads with parallel
  grippers,'' in {\em 2020 IEEE/RSJ International Conference on Intelligent
  Robots and Systems (IROS)}, pp.~2624--2629, IEEE, 2020.

\bibitem{takahashi2021target}
K.~Takahashi, N.~Fukaya, and A.~Ummadisingu, ``Target-mass grasping of
  entangled food using pre-grasping \& post-grasping,'' {\em IEEE Robotics and
  Automation Letters}, 2021.

\bibitem{moreira2016assessment}
E.~Moreira, L.~F. Rocha, A.~M. Pinto, A.~P. Moreira, and G.~Veiga, ``Assessment
  of robotic picking operations using a 6 axis force/torque sensor,'' {\em IEEE
  Robotics and Automation Letters}, vol.~1, no.~2, pp.~768--775, 2016.

\bibitem{hegemann2022learning}
P.~Hegemann, T.~Zechmeister, M.~Grotz, K.~Hitzler, and T.~Asfour, ``Learning
  symbolic failure detection for grasping and mobile manipulation tasks,'' in
  {\em 2022 IEEE/RSJ International Conference on Intelligent Robots and Systems
  (IROS)}, pp.~4302--4309, IEEE, 2022.

\end{thebibliography}
\end{document}